\pdfoutput=1

\documentclass[11pt]{article}

\usepackage{ACL2023}

\usepackage{times}
\usepackage{latexsym}

\usepackage[T1]{fontenc}

\usepackage[utf8]{inputenc}

\usepackage{microtype}

\usepackage{inconsolata}

\usepackage{multirow}
\usepackage{amsmath}
\DeclareMathOperator*{\argmax}{arg\,max}

\usepackage{amsfonts}

\usepackage{algorithm} 
\usepackage{algpseudocode}

\usepackage{booktabs}
\usepackage{siunitx}

\usepackage{marvosym}

\title{Supervising the Centroid Baseline \\for Extractive Multi-Document Summarization}

\author{Simão Gonçalves\textsuperscript{\Moon} \quad
        Gonçalo Correia\textsuperscript{\Moon} \quad
        Diogo Pernes\textsuperscript{\Moon\Saturn} \quad
        Afonso Mendes\textsuperscript{\Moon}\\
\textsuperscript{\Moon}Priberam Labs, Alameda D.\ Afonso Henriques, 41, 2º, 1000-123 Lisboa, Portugal\\
\textsuperscript{\Saturn}Faculdade de Engenharia da Universidade do Porto, Porto, Portugal \\
\{\href{mailto:simao.goncalves@priberam.pt}{\tt simao.goncalves},
\href{mailto:goncalo.correia@priberam.pt}{\tt goncalo.correia},
\href{mailto:diogo.pernes@priberam.pt}{\tt diogo.pernes}, \href{mailto:amm@priberam.pt}{\tt amm}\}{\tt @priberam.pt}
}

\begin{document}
\maketitle
\begin{abstract}
The centroid method is a simple approach for extractive multi-document summarization and many improvements to its pipeline have been proposed. We further refine it by adding a beam search process to the sentence selection and also a centroid estimation attention model that leads to improved results. We demonstrate this in several multi-document summarization datasets, including in a multilingual scenario.
\end{abstract}

\section{Introduction}


Multi-document summarization (MDS) addresses the need to condense content from multiple source documents into concise and coherent summaries while preserving the essential context and meaning.
Abstractive techniques, which involve generating novel text to summarize source documents, have gained traction in recent years \cite{liu-lapata-2019-hierarchical,jin-etal-2020-multi,xiao-etal-2022-primera}, following the advent of large pre-trained generative transformers. However, their effectiveness in summarizing multiple documents remains challenged. This is attributed not only to the long input context imposed by multiple documents but also to a notable susceptibility to factual inconsistencies. In abstractive methods, this is more pronounced when compared to their extractive counterparts due to the hallucination-proneness of large language models.

Extractive approaches, on the other hand, tackle this problem by identifying and selecting the most important sentences or passages from the given documents to construct a coherent summary. 
Extractive MDS usually involves a sentence importance estimation step \cite{hong-nenkova-2014-improving,cao2015ranking,cho-etal-2019-multi}, in which sentences from the source document are scored according to their relevance and redundancy with respect to the remaining sentences. Then, the summary is built by selecting a set of sentences achieving high relevance and low redundancy. The centroid-based method \cite{radev-etal-2000-centroid} is a cheap unsupervised solution in which each cluster of documents is represented by a centroid that consists of the sum of the TF-IDF representations of all the sentences within the cluster and the sentences are ranked by their cosine similarity to the centroid vector. While the original method is a baseline that can be easily surpassed, subsequent enhancements have been introduced to make it a more competitive yet simple approach \cite{rossiello-etal-2017-centroid,gholipour-ghalandari-2017-revisiting,lamsiyah2021unsupervised}.

In this work, we refine the centroid method even further: i) we utilize multilingual sentence embeddings to enable summarization of clusters of documents in various languages; ii) we employ beam search for sentence selection, leading to a more exhaustive exploration of the candidate space and ultimately enhancing summary quality; iii) we leverage recently proposed large datasets for multi-document summarization by adding supervision to the centroid estimation process. To achieve this, we train an attention-based model to approximate the oracle centroid obtained from the ground-truth target summary, leading to significant ROUGE-score improvements in mono and multilingual settings. To the best of our knowledge, we are the first to tackle the problem within a truly multilingual framework, enabling the summarization of a cluster of documents in different languages.\footnote{\url{https://github.com/Priberam/cera-summ}}


\section{Related Work}
\label{sec:related_work}
Typical supervised methods for extractive summarization involve training a model to predict sentence saliency, i.e.\ a model learns to score sentences in a document with respect to the target summary, either by direct match in case an extractive target is available or constructed \cite{svore-etal-2007-enhancing, woodsend-lapata-2012-multiple, mendes-etal-2019-jointly} or by maximizing a similarity score (\textit{e.g.}, ROUGE) with respect to the abstractive target summaries \cite{narayan-etal-2018-ranking}. Attempts to reduce redundancy exploit the notion of maximum marginal relevance \cite[MMR;][]{carbonell1998use, mcdonald2007study} or are coverage-based \cite{gillick2008icsi,almeida-martins-2013-fast}, seeking a set of sentences that cover as many concepts as possible while respecting a predefined budget. During inference, the model is then able to classify the sentences with respect to their salience, selecting the highest-scored sentences for the predicted summary.
Rather than training a model that predicts salience for each individual sentence, we employ a supervised model that directly predicts an overarching summary representation, specifically predicting the centroid vector of the desired summary.
Training this model can thus be more direct when training with abstractive summaries (as is the case in most summarization datasets), since computing the reference summary centroid is independent of whether the target is extractive or abstractive.

Regarding enhancements to the centroid method for extractive MDS, \citet{rossiello-etal-2017-centroid} refined it by substituting the TF-IDF representations with word2vec embeddings \citep{mikolov2013distributed}, and further incorporated a redundancy filter into the algorithm. \citet{gholipour-ghalandari-2017-revisiting}, on the other hand, retained the utilization of TF-IDF sentence representations but improved the sentence selection process. Recently, \citet{lamsiyah2021unsupervised} introduced modifications to the sentence scoring mechanism, incorporating novelty and position scores, and evaluated a diverse array of sentence embeddings with the proposed methodology, including contextual embeddings provided by ELMo \cite{peters-etal-2018-deep} and BERT \cite{devlin-etal-2019-bert}.

While there have been initiatives to foster research in multilingual extractive MDS \cite{giannakopoulos-2013-multi, giannakopoulos-etal-2015-multiling}, the proposed approaches \cite{litvak-vanetik-2013-multilingual, aries-etal-2015-allsummarizer, huang-2016-multilingual} are only language-agnostic, requiring all the documents within each cluster to be in the same language.  In contrast, we address extractive MDS in a scenario where each cluster is multilingual.

\section{Methodology}

The pipeline of our proposed model is divided into two stages. In the first stage, we use an attention model to obtain a cluster representation that replaces the naive centroid obtained by averaging sentence embeddings of the documents in a cluster. The rationale behind this approach is that the contribution of each sentence to the cluster centroid should depend on its relevance to the cluster summary. In order to capture the whole cluster context, a sentence-level attention model is employed, assigning variable weights to each sentence embedding so as to approximate the resulting average to the centroid that would be obtained by averaging the sentence embeddings of the target summary. In the second stage, an adapted version of the greedy sentence selection algorithm from \citet{gholipour-ghalandari-2017-revisiting} for extractive MDS is used to select the sentences included in the predicted summary. This adapted version uses our proposed supervised centroid and also includes a beam search algorithm to better explore the space of candidate summaries.

\subsection{Centroid Estimation}
\label{sec:centroid_estimation}
\citet{gholipour-ghalandari-2017-revisiting} builds a centroid by summing TF-IDF sentence representations of all the sentences that compose the cluster to summarize. In our research, we compute the centroid from a learnable weighted average of the contextual sentence embeddings, via an attention model.

\paragraph{Attention Model}
In our centroid estimation procedure, we use a pre-trained multilingual sentence transformer from \citet{yang-etal-2020-multilingual} to encode the sentences from the news articles, obtaining contextual embeddings $\boldsymbol{e}_k \in \mathbb{R}^{d}$, $k \in \{1, \dots, N\}$, for each of the $N$ sentences in a cluster.
Since it is often the case that the first sentences of a document are especially important for news summarization tasks,
we add sentence-level learnable positional embeddings to the contextual embeddings at the input of the attention model. Specifically, given a cluster $D$ comprising $N$ sentences, we compute:
\begin{equation}
    \boldsymbol{e}_{\text{pos}, k} = \boldsymbol{e}_k + \boldsymbol{p}_{\text{pos}(k)},
\end{equation}
where $\text{pos}(k)$ is the position within the respective document of the $k$-th sentence in the cluster and $\boldsymbol{p}_{\text{pos}(k)} \in \mathbb{R}^d$ is the corresponding learnable positional embedding. Each $\boldsymbol{e}_{\text{pos}, k} \in \mathbb{R}^{d}$ is then concatenated with the mean-pool vector of the cluster,\footnote{This is calculated by averaging the sentence embeddings within each document and then computing the mean of these individual document averages.} 
denoted by $\overline{\boldsymbol{e}_{\text{pos}}} \in\mathbb{R}^d$, resulting in $\boldsymbol{e}'_{\text{pos},k} = \mathrm{concat}(\boldsymbol{e}_{\text{pos},k},\overline{\boldsymbol{e}_{\text{pos}}})$
for each sentence. This concatenation ensures that the computation of the attention weight for each position uses information from all the remaining positions.
The vector $\boldsymbol{\beta} \in \mathbb{R}^{N}$ of attention weights is obtained as:
\begin{equation}
    \boldsymbol{\beta} = \mathrm{softmax}\left(\text{MLP}(\boldsymbol{e}'_{\text{pos},1}), \dots, \text{MLP}(\boldsymbol{e}'_{\text{pos},N}) \right),
\end{equation}
where MLP is a two-layer perceptron shared by all the positions. It has a single output neuron and a hidden layer with $d$ units and a tanh activation. 

After computing the attention weights for the cluster, we take the original sentence embeddings $\boldsymbol{e}_k$, $k \in \{1, \dots, N\}$, and compute a weighted sum of these representations:
\begin{equation}
\label{eq:attention_estimation}
    \boldsymbol{h} = \sum_{k=1}^{N}{\beta_{k}\boldsymbol{e}_k}.
\end{equation}
Consequently, the resultant vector $\boldsymbol{h} \in \mathbb{R}^d$ is a convex combination of the input sentence embeddings. Since it is not guaranteed that the target centroid lies within this space, $\boldsymbol{h}$ is subsequently mapped to the output space through a linear layer, yielding an estimate $\hat{\boldsymbol{c}}_\text{attn} \in \mathbb{R}^d$ of the centroid.
Hereafter we refer to this attention model as \textbf{Ce}ntroid \textbf{R}egression \textbf{A}ttention (CeRA).
\paragraph{Interpolation}
The original (unsupervised) approach involves estimating the centroid by computing the average of all sentence representations $\boldsymbol{e}_{k}$ within a cluster, which has consistently demonstrated strong performance. Let $\overline{\boldsymbol{e}}_D$ represent this centroid for cluster $D$. To leverage the advantages of this effective technique, we introduce $\overline{\boldsymbol{e}}_D$ as a residual component to enhance the estimate produced by the attention model. Thus, our final centroid estimate is computed as:
\begin{equation}
\label{interpol}
    \hat{\boldsymbol{c}} = \boldsymbol{\alpha} \odot \hat{\boldsymbol{c}}_\text{attn} + (1-\boldsymbol{\alpha}) \odot \overline{\boldsymbol{e}}_D,
\end{equation}
where $\boldsymbol{\alpha} \in [0,1]^d$ is a vector of interpolation weights and $\odot$ denotes elementwise multiplication. The interpolation weights are obtained from concatenating $\hat{\boldsymbol{c}}_\text{attn}$ and $\overline{\boldsymbol{e}}_D$ and mapping it through an MLP of two linear linear layers with $d$ units each. The two layers are interleaved with a ReLU activation and a sigmoid is applied at the output. We call the model with interpolation CeRAI.

\paragraph{Training Objective}
Finally, we minimize the cosine distance between the model predictions $\hat{\boldsymbol{c}}$ and the mean-pool of the sentence embeddings of the target summary $\boldsymbol{c}_{\text{gold}}$.
\subsection{Sentence Selection}
\label{sec:sentence_selection}
Considering the cluster $D$ and a set $S$ with the current sentences in the summary. at each iteration of greedy sentence selection \citep{gholipour-ghalandari-2017-revisiting}, we have
\begin{equation}
    \boldsymbol{e}_{S\cup{\{s\}}} = \sum_{s' \in S} \boldsymbol{e}_{s'} + \boldsymbol{e}_{s}
\end{equation}
for each sentence $s \in D \setminus S$. Then, the new sentence $s^*$ to be included in the summary is
\begin{equation}
    \label{eq:sentence_scores}
    s^* = \argmax_{s \in D \setminus S} \cos \mathrm{sim}(\boldsymbol{e}_{S\cup{\{s\}}} ,\overline{\boldsymbol{e}}_D),
\end{equation}
where $\cos \mathrm{sim}$ is the cosine similarity. The algorithm stops when the summary length reaches the specified budget.\footnote{While the original algorithm would stop after the first sentence that exceeded the budget, we stop before it is exceeded, and thus we do not need truncation to respect the budget.}
As demonstrated in that work, redundancy is mitigated since the centroid is compared to the whole candidate summary $S\cup{\{s\}}$ at each iteration and not only to the new sentence $s$. 
 
In our version of the algorithm, we not only estimate the cluster centroids as explained in §\ref{sec:centroid_estimation}, replacing $\overline{\boldsymbol{e}}_D$ by $\hat{\boldsymbol{c}}$ in equation~\eqref{eq:sentence_scores}, but also employ a beam search (BS) algorithm so that the space of candidate summaries is explored more thoroughly. Moreover, in order to exhaust the chosen budget, we add a final greedy search to do further improvements to the extracted summary. The procedure is defined in Algorithm~\ref{alg:sum_alg}, shown in Appendix~\ref{app:sentence_selection_algorithm}, and we describe it less formally below.

\begin{table*}[htbp]
\centering
\small
\begin{tabular}{lrrrr} \toprule
        Method
        & Multi-News
        & WCEP-10
        & TAC2008
        & DUC2004 \\ \midrule
        Oracle centroid 
        & 21.72 {\tiny\color{gray}$\pm$ 0.33} 
        & 28.54 {\tiny\color{gray}$\pm$ 1.21} 
        & 11.99 {\tiny\color{gray}$\pm$ 1.32} 
        & 10.29{\tiny\color{gray}$\pm$ 1.01} 
        \\ \midrule
    \citeauthor{gholipour-ghalandari-2017-revisiting}  
        & 16.07 {\tiny\color{gray}$\pm$ 0.26} 
        & 15.09 {\tiny\color{gray}$\pm$ 0.92} 
        & 7.36 {\tiny\color{gray}$\pm$ 1.15} 
        & 6.82 {\tiny\color{gray}$\pm$ 0.76} 
        \\
    \citeauthor{lamsiyah2021unsupervised}  
        & 13.92 {\tiny\color{gray}$\pm$ 0.22} 
        & 16.10 {\tiny\color{gray}$\pm$ 0.96} 
        & 7.91 {\tiny\color{gray}$\pm$ 1.31} 
        & \textbf{7.80} {\tiny\color{gray}$\pm$ 0.78}
        \\
    BS (\textit{Ours})  
        & 16.22 {\tiny\color{gray}$\pm$ 0.25} 
        & 15.64 {\tiny\color{gray}$\pm$ 0.97} 
        & 8.10 {\tiny\color{gray}$\pm$ 1.32} 
        & 7.03 {\tiny\color{gray}$\pm$ 0.64} 
        \\
    BS+GS (\textit{Ours})  
        & 16.70 {\tiny\color{gray}$\pm$ 0.26} 
        & 16.41 {\tiny\color{gray}$\pm$ 0.91} 
        & 8.16 {\tiny\color{gray}$\pm$ 1.25} 
        & 7.46 {\tiny\color{gray}$\pm$ 0.83} 
        \\
    CeRA (\textit{Ours})
        & 17.98 {\tiny\color{gray}$\pm$ 0.23} 
        & \textbf{17.46} {\tiny\color{gray}$\pm$ 0.98} 
        & 8.27 {\tiny\color{gray}$\pm$ 1.26} 
        & 7.31 {\tiny\color{gray}$\pm$ 0.74} 
        \\
    CeRAI (\textit{Ours})
        & \textbf{17.99} {\tiny\color{gray}$\pm$ 0.27} 
        & 17.24 {\tiny\color{gray}$\pm$ 0.93} 
        & \textbf{8.37} {\tiny\color{gray}$\pm$ 1.24} 
        & 7.72 {\tiny\color{gray}$\pm$ 0.77} 
        \\
    \bottomrule
\end{tabular}
\caption{ROUGE-2 recall with 95\% bootstrap confidence intervals of different extractive methods on the considered test sets. CeRA and CeRAI were only trained on the Multi-News training dataset.
}
\label{res:oursys}
\end{table*}

\paragraph{Beam Search}
The process begins by pre-selecting sentences, retaining only the first $n$ sentences from each document. Beam search initiates by selecting the top $B$ sentences with the highest similarity scores with the centroid, where $B$ represents the beam size. In each subsequent iteration, the algorithm finds the highest-scoring $B$ sentences on each beam, generating a total of $B^2$ candidates. Among these candidates, only the highest-ranked $B$ sentences are retained. Suppose any of these sentences exceed the specified budget length for the summary. In that case, we preserve the corresponding previous state, and no further exploration is conducted on that beam. The beam search concludes when all candidate beams have exceeded the budget or when no more sentences are available.


\paragraph{Greedy Search}
To exhaust the specified budget and improve results, we add a greedy search of sentences that are allowed within the word limit. The top-scoring $B$ states from the beam search are used as starting points for this greedy search. Then, for each state, we greedily select the highest-scoring sentence that does not exceed the budget among the top $T$ ranked sentences. This process iterates until either all of the top $T$ ranked sentences would exceed the budget or there are no further sentences left for consideration.

\section{Experimental Setup}

Herein, we outline the methods, datasets, and evaluation metrics employed in our experiments.

\paragraph{Methods}
We compare our approaches with the centroid-based methods from \citet{gholipour-ghalandari-2017-revisiting} and \citet{lamsiyah2021unsupervised}, described in §\ref{sec:related_work}. To be consistent with the remaining methods, the approach by \citet{gholipour-ghalandari-2017-revisiting} was implemented on top of contextual sentence embeddings instead of TF-IDF. Additionally, we perform ablation evaluations in three scenarios: i) a scenario (BS) where we do not use the centroid estimation model (§\ref{sec:centroid_estimation}) and rely solely on the beam search for the sentence selection step (§\ref{sec:sentence_selection}); ii) a scenario (BS+GS) identical to the previous one, except that we perform the greedy search step after the beam search; iii) two scenarios (CeRAI and CeRA) where we utilize the centroid estimation model with and without incorporating interpolation, and apply the BS+GS algorithm on the predicted centroid. The ``Oracle centroid'' upperbounds our approaches, since it results from applying BS+GS on the mean-pool of the sentence embeddings of the target summary, $\boldsymbol{c}_{\text{gold}}$, as the cluster centroid.
Appendix~\ref{app:experimental_details} provides additional details about data processing and hyperparameters.

\paragraph{Datasets}
We used four English datasets, Multi-News \cite{fabbri-etal-2019-multi}, WCEP-10 \cite{gholipour-ghalandari-etal-2020-large}, TAC2008, and DUC2004, and one multilingual dataset, CrossSum \cite{bhattacharjee-etal-2023-crosssum}, in our experiments. We used the centroid-estimation models trained on Multi-News to evaluate CeRA and CeRAI on WCEP-10, TAC2008, and DUC2004 since these datasets do not provide training splits. CrossSum was conceived for single-document cross-lingual summarization, so we had to adapt it for multilingual MDS. This adaptation results in clusters that encompass documents in multiple languages, with each cluster being associated with a single reference summary containing sentences in various languages. We explain this procedure and provide further details about each dataset in Appendix~\ref{app:datasets}.

\paragraph{Evaluation Metrics}
We evaluate ROUGE scores \cite{lin-2004-rouge} in all the experiments. When evaluating models in the multilingual setting, we translated both the reference summaries and the extracted summaries into English prior to ROUGE computation. As we optimized for R2-R on the validation sets, we report it as our main metric in Tables~\ref{res:oursys} and \ref{res:crosssum}. The remaining scores are shown in Appendix~\ref{app:additional_results}.

\section{Results}

\paragraph{Monolingual Setting}
The ROUGE-2 recall (R2-R) of all the methods in the monolingual datasets are presented in Table~\ref{res:oursys}. F1 scores and results for the other ROUGE variants are presented in Table~\ref{res:additional_mono}, in Appendix~\ref{app:additional_results}. The first observation is that BS alone outperforms \citet{gholipour-ghalandari-2017-revisiting} in all datasets, with additional improvements obtained when the greedy search step is also performed (BS+GD). This was expected since our approach explores the candidate space more thoroughly. The motivation for using a supervised centroid estimation model arose from the excellent ROUGE results obtained when using the target summaries to build the centroid (``Oracle centroid'' in the tables), showing that an enhanced centroid estimation procedure could improve the results substantially. This is confirmed by the two methods using the centroid estimation model (CeRA and CeRAI), which improve R2-R significantly in Multi-News and WCEP-10 and perform at least on par with \citet{lamsiyah2021unsupervised} in TAC2008 and DUC2004. 
It's also worth noting that CeRA and CeRAI were only trained on the Multi-News training set and nevertheless performed better or on par with the remaining baselines on the test sets of the remaining corpora.
Incorporating the interpolation step (CeRAI) appears to yield supplementary enhancements compared to the non-interpolated version (CeRA) across various settings, which we attribute to this method adding regularization to the estimation process, improving results on harder scenarios.

\begin{table}[t]
\centering
\small
\begin{tabular}{lrr} \toprule
    Method 
        & CrossSum 
        & CrossSum-ZS  \\ \midrule
    Oracle centroid  
        & 11.74 {\tiny\color{gray}$\pm$ 0.55}
        & 14.91 {\tiny\color{gray}$\pm$ 0.49}
        \\ \midrule
    \citeauthor{gholipour-ghalandari-2017-revisiting}  
        & 7.72 {\tiny\color{gray}$\pm$ 0.43}
        & 10.03 {\tiny\color{gray}$\pm$ 0.40}
        \\
    \citeauthor{lamsiyah2021unsupervised}  
        &  8.01 {\tiny\color{gray}$\pm$ 0.52}
        & 10.45 {\tiny\color{gray}$\pm$ 0.46}
        \\
    BS (\textit{Ours})  
        & 7.74 {\tiny\color{gray}$\pm$ 0.44}
        & 10.16 {\tiny\color{gray}$\pm$ 0.40}
        \\
    BS+GS (\textit{Ours})  
        & 8.23 {\tiny\color{gray}$\pm$ 0.43}
        & 10.85 {\tiny\color{gray}$\pm$ 0.41}
        \\
    CeRA (\textit{Ours})
        & \textbf{9.65} {\tiny\color{gray}$\pm$ 0.49}
        & 11.67 {\tiny\color{gray}$\pm$ 0.41}
        \\
    CeRAI (\textit{Ours})
        & 9.38 {\tiny\color{gray}$\pm$ 0.50}
        & \textbf{11.73} {\tiny\color{gray}$\pm$ 0.43}
        \\
    \bottomrule
\end{tabular}
\caption{ROUGE-2 recall results with 95\% bootstrap confidence intervals of different extractive methods on the multilingual test sets. The CrossSum set contains the same languages used for training the centroid estimation model, whereas CrossSum-ZS (\textit{zero-shot}) consists of languages that were not present in the training data.}
\label{res:crosssum}
\end{table}

\paragraph{Multilingual Setting}
The R2-R scores of all the methods in CrossSum can be found in Table~\ref{res:crosssum}, while additional results are in Table~\ref{res:additional_multi} of Appendix~\ref{app:additional_results}. Once again, we observe the superiority of the centroid estimation models, CeRA and CeRAI, in comparison to all the remaining methods, with the variants with and without interpolation performing on par with each other. Most notably, these models prove to be useful even when tested with languages unseen during the training phase, underscoring their robustness and applicability in a zero-shot setting.


\section{Conclusions}


We enhanced the centroid method for multi-document summarization by extending a previous approach with a beam search followed by a greedy search. Additionally, we introduced a novel attention-based regression model for better centroid prediction. These improvements outperform existing methods across various datasets, including a multilingual setting, offering a robust solution for this challenging scenario. Regarding future work, we believe an interesting research direction would be to further explore using the supervised centroids obtained by the CeRA and CeRAI models, by having them as a proxy objective to obtain improved abstractive summaries.

\section*{Limitations}
While we believe that our approach possesses merits, it is equally important to recognize its inherent limitations. Diverging from conventional centroid methods that operate entirely in an unsupervised manner, our centroid estimation model necessitates training with reference summaries. Nevertheless, its robustness to dataset shifts was demonstrated: the model trained on Multi-News consistently yielded strong results when assessed on different English datasets, and the model trained on a subset of languages from CrossSum displayed successful generalization to other languages.

Finally, our method introduces increased computational complexity. This arises from both the forward pass through the attention model and the proposed beam search algorithm, which incurs a greater computational cost compared to the original, simpler greedy approach proposed by \citet{gholipour-ghalandari-2017-revisiting}.

\section*{Acknowledgements}
This work is supported by the EU H2020 SELMA project (grant agreement No.\ 957017).

\bibliography{anthology,custom}
\bibliographystyle{acl_natbib}

\newpage

\appendix

\section{Sentence Selection Algorithm}
\label{app:sentence_selection_algorithm}

\begin{algorithm}
    \caption{Sentence Selection} 
    \label{alg:sum_alg}
    
    \begin{algorithmic}[1]
        \Require Cluster $D$, centroid $\hat{\boldsymbol{c}}$, summary budget $\ell$, number of  sentences $n$ to pre-select, beam size $B$, number of candidates $T$ for greedy search.
        \State $D_{n} \gets \text{select-first}(D,n)$  
        \State$\pi, \pi_{\text{next}}, \pi_{\text{bs}} \gets \text{empty list}$
        
	\While {$\exists b: \text{length}(\pi_{\text{next}}[b]) < \ell$}:  \Comment{Beam Search}
            \State $\pi_{\text{next}} \gets \text{BS}_{\text{step}}(\pi,D_{n},B,\hat{\boldsymbol{c}})$ $^($\footnotemark${^)}$
            \If{$\exists b: \text{length}(\pi_{\text{next}}[b]) > \ell$}
		      \State $\pi_{\text{bs}}.\text{append} (\pi)$ 
            \EndIf
            \State $\pi \gets \forall \pi_{\text{next}}[b]: \text{length}(\pi_{\text{next}}[b]) \leq \ell$
        \EndWhile
        
        \State $\pi_{\text{best}} \gets$ highest-scored $B$ states in $\pi_{\text{bs}}$ (sorted)
        \For{$b = 1, 2, \dots, B$}: \Comment{Greedy Search}
            \State $t \gets 0$
            \State $D'_{n} \gets D_{n} \setminus  \pi_{\text{best}}[b]$
            \While{$t < T$}:
                \State $s^* \gets   \argmax\limits_{s \in D'_{n}} \cos \mathrm{sim}(\boldsymbol{e}_{\pi_{\text{best}}[b]\cup{\{s\}}}, \hat{\boldsymbol{c}})$
                
                \State $\pi'_{\text{best}}[b] \gets \pi_{\text{best}}[b] \cup \{s^*\}$
                \If{$\text{length}(\pi'_{\text{best}}[b]) \leq \ell$}:
                    \State $\pi_{\text{best}}[b] \gets \pi'_{\text{best}}[b]$ 
                    \State $t \gets 0$
                \Else:
                    \State $t \gets t+1$
                \EndIf
                \State $D'_{n} \gets D'_{n} \setminus \{s^*\} $
            \EndWhile
        \EndFor    
        \State \Return $S \gets$ highest-scored state in $\pi_{\text{best}}$
   \end{algorithmic}
\end{algorithm}
\footnotetext{$\text{BS}_{\text{step}}$ denotes a step of the usual beam search algorithm. Details omitted for brevity.}

\section{Datasets}
\label{app:datasets}

We now describe each of the datasets used for evaluation and explain how we have adapted CrossSum for the task of MDS.

\paragraph{Multi-News} The Multi-News dataset \cite{fabbri-etal-2019-multi} is a large-scale dataset for MDS of news articles. It contains up to 10 documents per cluster and more than 50 thousand clusters divided into training, validation, and test splits. There is a single human-written reference summary for each cluster.

\paragraph{WCEP-10} This dataset \citep{DBLP:journals/corr/abs-2005-10070,xiao-etal-2022-primera} consists of short human-written target summaries extracted from the Wikipedia Current Events Portal (WCEP). Each news cluster associated with a certain event is paired with a single reference summary, and there are at most 10 documents per cluster. The dataset comprises 1022 clusters, all of which are used for testing.

\paragraph{TAC2008} This is a multi-reference dataset introduced by the Text Analysis Conference (TAC)\footnote{\url{https://tac.nist.gov}}. It provides no training nor validation sets and the test set consists of 48 news clusters, each with 10 related documents and 4 human-written summaries as references.

\paragraph{DUC2004} Another multi-reference news summarization dataset\footnote{\url{https://duc.nist.gov}} designed and used for testing only. It contains 50 clusters with 10 documents and 4 human-written reference summaries each.

\paragraph{CrossSum} To assess the performance of the models in a multilingual context, we have adapted the CrossSum dataset \cite{bhattacharjee-etal-2023-crosssum} for the task of MDS. Initially designed for cross-lingual summarization, this dataset offers document-summary pairs for more than 1500 language directions. The dataset is derived from pairs of articles sourced from the multilingual summarization dataset XL-Sum \cite{hasan-etal-2021-xl}. Notably, these pairings were established using an automatic similarity metric, resulting in many pairs covering similar topics rather than the exact same stories, rendering it well-suited MDS.

To tailor this dataset for our specific task, we began by selecting the data from a predefined subset of the languages. Subsequently, we aggregated the documents into clusters, taking into account their pairings. For instance, if document $A$ was paired with document $B$ and document $B$ was paired with document $C$, then $A$, $B$, and $C$ would belong to the same cluster. Clusters containing only one document were discarded. For obtaining multilingual reference summaries for each cluster, we interleaved the sentences from the individual summaries until we reached a predefined limit of 100 words. We have built training, validation, and test sets using data in English, Spanish, and French, and another test set using data in Portuguese, Russian, and Turkish to evaluate our model in a zero-shot setting. Statistics about each split are presented in Table~\ref{tab:crosssum_stats}.

\section{Experimental Details}
\label{app:experimental_details}

\paragraph{Data Processing}
To ensure a fair comparison, all the models we evaluated used the same sentence representations, specifically, sentence embeddings obtained from the \texttt{distiluse-base-multilingual-cased-v2}\footnote{\url{https://huggingface.co/sentence-transformers/distiluse-base-multilingual-cased-v2}} sentence encoder \cite{yang-etal-2020-multilingual}.

For monolingual datasets, the documents were split into sentences using \texttt{sent\_tokenize} from the NLTK library \cite{bird2009natural}. For CrossSum, we used \texttt{SentSplitter} from the multilingual ICU-tokenizer.\footnote{\url{https://pypi.org/project/icu-tokenizer}} Regular expressions were applied to replace redundant white spaces and excessive paragraphs and empty sentences were excluded. Before sentence selection (Algorithm  \ref{alg:sum_alg}), the data goes through a second processing step, during which duplicate sentences and sentences that individually exceed the summary budget are eliminated.

When evaluating models in CrossSum, we translated both the reference summaries and the extracted summaries into English prior to ROUGE computation. All the translations were performed using the M2M-100 12-billion-parameter model \cite{fan2021beyond}.

The following word-limit budgets were used by all models: 230 words for the Multi-News dataset, 100 words for TAC2008, DUC2004 and CrossSum, and 50 words for WCEP-10.\footnote{We used ROUGE 1.5.5 toolkit with the following arguments: \texttt{-n 4 -m -2 4 -l budget -u -c 95 -r 1000 -f A -p 0.5 -t 0 -a}}

\paragraph{Hyperparameters} The hyperparameters for the beam search-based methods were tuned by running a grid search on the BS+GS approach on the Multi-News validation set. For the number of sentences $n$, odd numbers from 1 to 9 were tested. For the beam width $B$ values 1,5, and 9 were examined, and regarding the number of candidates $T$, values 1,5, and 9 were considered. The values that maximized R2-R on this validation set were $n$ = 9, $B$=5, and $T$=9. In all of our experiments, these were the values we considered for the parameters. Note that for the BS method only $n$ and $B$ are relevant.\par
The hyperparameters of the centroid estimation model used in CeRA were obtained by random search on Multi-News. The hyperparameters yielding the highest R2-R score on the validation set for the produced summaries were kept. The CeRAI model was trained using the optimal hyperparameters found for CeRA. The optimal parameters were: \textit{batch size} = 2, \textit{learning rate} = 5$\times$10$^{-4}$, and \textit{number of positional encodings} = 35. We utilized the Adam optimizer with a multi-step learning rate scheduler configured with \textit{step size} = 3 and $\gamma$ = 0.1.

\paragraph{Implementation Details} Our CeRA and CeRAI models used early stopping, where the stopping criteria metric was based on R2-R. Layer normalization \cite{ba2016layer} was applied on the input data before adding the positional information to it and before passing the data through the last linear layer that transforms $\boldsymbol{h}$ (equation~(\ref{eq:attention_estimation})) into $\hat{\boldsymbol{c}}_\text{attn}$ in the CeRA and CeRAI models. We have also normalized the input data to have a unit L2 norm.

\begin{table*}[htbp]
\centering
\small
\begin{tabular}{lrrrrr} \toprule
Split   & Languages & \#Clusters & \#Docs per cluster & Avg \#sentences per doc & Avg \#words per summary \\ \midrule
Train   & en, es, fr                & 6541                          & 2$-$10 & 38.5 {\tiny\color{gray}$\pm$ 28.8} & 52.5 {\tiny\color{gray}$\pm$ 16.1}                                  \\
Val     & en, es, fr                & 889                           & 2$-$6 & 34.4 {\tiny\color{gray}$\pm$ 27.4} & 52.3 {\tiny\color{gray}$\pm$ 15.5}                                  \\
Test    & en, es, fr                & 853                           & 2$-$6 & 36.6 {\tiny\color{gray}$\pm$ 35.4} & 52.2 {\tiny\color{gray}$\pm$ 16.2}                                  \\
Test-ZS & pt, ru, tr                & 933                          & 2$-$5 & 23.4 {\tiny\color{gray}$\pm$ 21.1} & 60.2 {\tiny\color{gray}$\pm$ 20.8}  \\                     \bottomrule         
\end{tabular}
\caption{CrossSum: statistics of each split. Averages are indicated with standard deviations.}
\label{tab:crosssum_stats}
\end{table*}

\section{Additional Results}
\label{app:additional_results}
The ROUGE-1/2/L recall and F1 scores obtained by all the methods in the monolingual datasets are shown in Table~\ref{res:additional_mono}. Table~\ref{res:additional_multi} presents the same quantities for the multilingual case.

\begin{table*}[htbp]
\centering
\small
\begin{tabular}{llrrrrrr} \toprule
     
     Test set & Method   & R1-R & R1-F & R2-R & R2-F & RL-R & RL-F \\ \midrule
    \multirow{6}{*}{Multi-News} & Oracle centroid
        & 54.26 & 50.36
        & 21.72 & 20.02
        & 24.33 & 22.42
        \\ 
    &\citeauthor{gholipour-ghalandari-2017-revisiting}  
        & 47.91 & 45.64 
        & 16.07 & 15.16
        & 21.41 & 20.24 
        \\
     &\citeauthor{lamsiyah2021unsupervised}  
        & 44.91 & 43.02 
        & 13.93 & 13.18 
        & 20.56 & 19.53 
        \\
    & BS (\textit{Ours})  
        & 48.34 & 45.81 
        & 16.22 & 15.24 
        & 21.34 & 20.08 
        \\
    & BS+GS (\textit{Ours})  
        & 49.54 & 45.98 
        & 16.70 & 15.36 
        & 21.81 & 20.08 
        \\
    & CeRA (\textit{Ours})
        & 50.75 & 47.07 
        & 17.98 & 16.52 
        & \textbf{22.69} & 20.86 
        \\
    & CeRAI (\textit{Ours})
        & \textbf{50.76} & \textbf{47.08} 
        & \textbf{17.99} & \textbf{16.53} 
        & \textbf{22.69} & \textbf{20.87} 
        \\ \midrule
    \multirow{6}{*}{WCEP-10} & Oracle centroid
        & 58.72 & 44.94
        & 28.54 & 21.50
        & 42.38 & 31.94
        \\ 
     &\citeauthor{gholipour-ghalandari-2017-revisiting}  
        & 41.26 & 35.09 
        & 15.09 & 12.61 
        & 29.42 & 24.86 
        \\
     &\citeauthor{lamsiyah2021unsupervised}  
        & 41.65 & \textbf{35.62} 
        & 16.10 & \textbf{13.38} 
        & 30.53 & \textbf{25.75} 
        \\
    & BS (\textit{Ours})  
        & 43.48 & 35.07 
        & 15.64 & 12.42 
        & 30.49 & 24.44 
        \\
    & BS+GS (\textit{Ours})  
        & 46.23 & 34.72 
        & 16.41 & 12.05 
        & 31.85 & 23.60 
        \\
    & CeRA (\textit{Ours})
        & \textbf{47.14} & 35.23 
        & \textbf{17.46} & 12.65 
        & \textbf{33.03} & 24.28
        \\
    & CeRAI (\textit{Ours})
        & 46.85 & 35.17 
        & 17.24 & 12.59 
        & 32.81 & 24.24 
        \\
    \midrule
    \multirow{6}{*}{TAC2008} &  Oracle centroid
        & 41.07 & 42.02
        & 11.99 & 12.26
        & 20.66 & 21.11
        \\
     &\citeauthor{gholipour-ghalandari-2017-revisiting}  
        & 32.00 & 34.38
        &  7.36 &  7.91 
        & 16.64 & 17.87 
        \\
     &\citeauthor{lamsiyah2021unsupervised}  
        & 31.00 & 33.75 
        &  7.91 &  \textbf{8.65} 
        & 16.65 & 18.16 
        \\
    & BS (\textit{Ours})  
        & 33.93 & 35.62 
        &  8.10 &  8.53
        & 17.62 & \textbf{18.50}  
        \\
    & BS+GS (\textit{Ours})  
        & \textbf{35.12} & \textbf{35.98} 
        &  8.16 &  8.34 
        & \textbf{17.99} & 18.40 
        \\
    & CeRA (\textit{Ours})
        & 34.43 & 35.07 
        &  8.27 &  8.42 
        & 17.35 & 17.66 
        \\
    & CeRAI (\textit{Ours})
        & 34.44 & 35.11 
        &  \textbf{8.37} &  8.52 
        & 17.73 & 18.06 
        \\ \midrule
    \multirow{6}{*}{DUC2004} & Oracle centroid
        & 39.93 & 41.10
        & 10.29 & 10.60
        & 19.48 & 20.05
        \\
    &\citeauthor{gholipour-ghalandari-2017-revisiting}  
        & 32.82 & 35.86 
        &  6.82 &  7.48 
        & 16.00 & 17.51
        \\
     &\citeauthor{lamsiyah2021unsupervised}  
        & 32.81 & 36.03 
        &  \textbf{7.80} &  \textbf{8.61} 
        & 16.66 & \textbf{18.34} 
        \\
    & BS (\textit{Ours})  
        & 34.01 & 36.20 
        &  7.03 &  7.51 
        & 16.35 & 17.41 
        \\
    & BS+GS (\textit{Ours})
        & 35.11 & 36.37 
        &  7.46 &  7.74 
        & \textbf{16.98} & 17.60 
        \\
    & CeRA (\textit{Ours})
        & 34.88 & 36.06 
        &  7.31 &  7.56 
        & 16.67 & 17.23 
        \\
    & CeRAI (\textit{Ours})
        & \textbf{35.16} & \textbf{36.38} 
        &  7.72 &  7.99 
        & 16.89 & 17.48 
        \\
    \bottomrule
\end{tabular}
\caption{ROUGE-1/2/L recall and F1 results of different extractive methods on the considered monolingual test sets.}
\label{res:additional_mono}
\end{table*}

\begin{table*}[htbp]
\centering
\small
\begin{tabular}{llrrrrrr} \toprule
     
     Test set & Method   & R1-R & R1-F & R2-R & R2-F & RL-R & RL-F \\ \midrule
    \multirow{6}{*}{CrossSum} & Oracle centroid  
        & 46.86 & 31.85
        & 11.74 &  7.93
        & 27.64 & 18.57
        \\
    &\citeauthor{gholipour-ghalandari-2017-revisiting}  
        & 38.64 & 27.88 
        &  7.72 &  5.56
        & 23.30 & 16.65 
        \\
     &\citeauthor{lamsiyah2021unsupervised}  
        & 37.89 & 27.53 
        &  8.01 &  5.77 
        & 23.81 & 17.13 
        \\
    & BS (\textit{Ours})  
        & 39.24 & 27.83 
        &  7.74 &  5.48 
        & 23.60 & 16.53 
        \\
    & BS+GS (\textit{Ours})  
        & 40.78 & 27.71 
        &  8.23 &  5.57 
        & 24.42 & 16.39 
        \\
    & CeRA (\textit{Ours})
        & \textbf{42.45} & \textbf{28.89} 
        &  \textbf{9.65} &  \textbf{6.52} 
        & \textbf{25.64} & \textbf{17.27} 
        \\
    & CeRAI (\textit{Ours})
        & 42.31 & 28.73 
        &  9.38 &  6.31
        & 25.55 & 17.15 
        \\ \midrule
    \multirow{6}{*}{CrossSum-ZS} & Oracle centroid  
        & 50.55 & 37.30
        & 14.91 &  11.00
        & 28.90 & 21.08
        \\
    &\citeauthor{gholipour-ghalandari-2017-revisiting}  
        & 41.70 & 32.65 
        & 10.03 &  7.82 
        & 24.52 & 19.02 
        \\
     &\citeauthor{lamsiyah2021unsupervised}  
        & 41.14 & 32.39 
        & 10.45 &  8.17 
        & 24.81 & 19.31 
        \\
    & BS (\textit{Ours})  
        & 42.53 & 32.65 
        & 10.16 &  7.81
        & 24.87 & 18.90 
        \\
    & BS+GS (\textit{Ours})  
        & 44.36 & 32.65 
        & 10.85 &  7.99 
        & 25.74 & 18.74 
        \\
    & CeRA (\textit{Ours})
        & \textbf{45.44} & \textbf{33.43} 
        & 11.67 & 8.57 
        & \textbf{26.52} & \textbf{19.30}
        \\
    & CeRAI (\textit{Ours})
        & 45.37 & 33.38 
        & \textbf{11.73} &  \textbf{8.62} 
        & 26.51 & 19.26 
        \\
    \bottomrule
\end{tabular}
\caption{ROUGE-1/2/L recall and F1 results of different extractive methods on the considered multilingual test sets.}
\label{res:additional_multi}
\end{table*}

\end{document}